\documentclass[10pt,twocolumn,letterpaper]{article}

\usepackage{iccv}              % To produce the CAMERA-READY version
\definecolor{iccvblue}{rgb}{0.21,0.49,0.74}
\usepackage[pagebackref,breaklinks,colorlinks,allcolors=iccvblue]{hyperref}

\usepackage{tcolorbox}
\usepackage{pifont}
\usepackage{longtable}
\usepackage{makecell}
\usepackage{multirow} 
\usepackage{xcolor}
\usepackage[title]{appendix}
\usepackage[accsupp]{axessibility}
\usepackage{ulem}
\usepackage{graphicx}
\usepackage{fontawesome}

\def\paperID{272} % *** Enter the Paper ID here
\def\confName{ICCV}
\def\confYear{2025}

\title{\vspace{-1cm}\raisebox{-0.1cm}{\includegraphics[width=0.2\textwidth,height=0.03\textwidth]{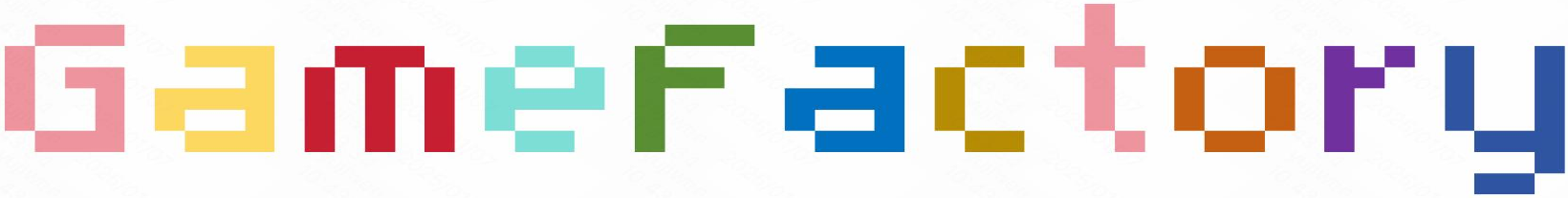}} : Creating New Games with Generative Interactive Videos}

\author{Jiwen Yu$^{1*\dagger}$\qquad Yiran Qin$^{1*}$ \qquad Xintao Wang$^{2\ddagger}$ \qquad  Pengfei Wan$^{2}$\qquad  Di Zhang$^{2}$\qquad  Xihui Liu$^{1\ddagger}$\\
\normalsize$^1$ The University of Hong Kong \qquad \normalsize$^2$ Kuaishou Technology 
\\
\url{https://yujiwen.github.io/gamefactory/}
}

\begin{document}
% \maketitle
\twocolumn[{
\renewcommand\twocolumn[1][]{#1}
\maketitle
\centering
\vspace{-0.6cm}
\includegraphics[width=\textwidth]{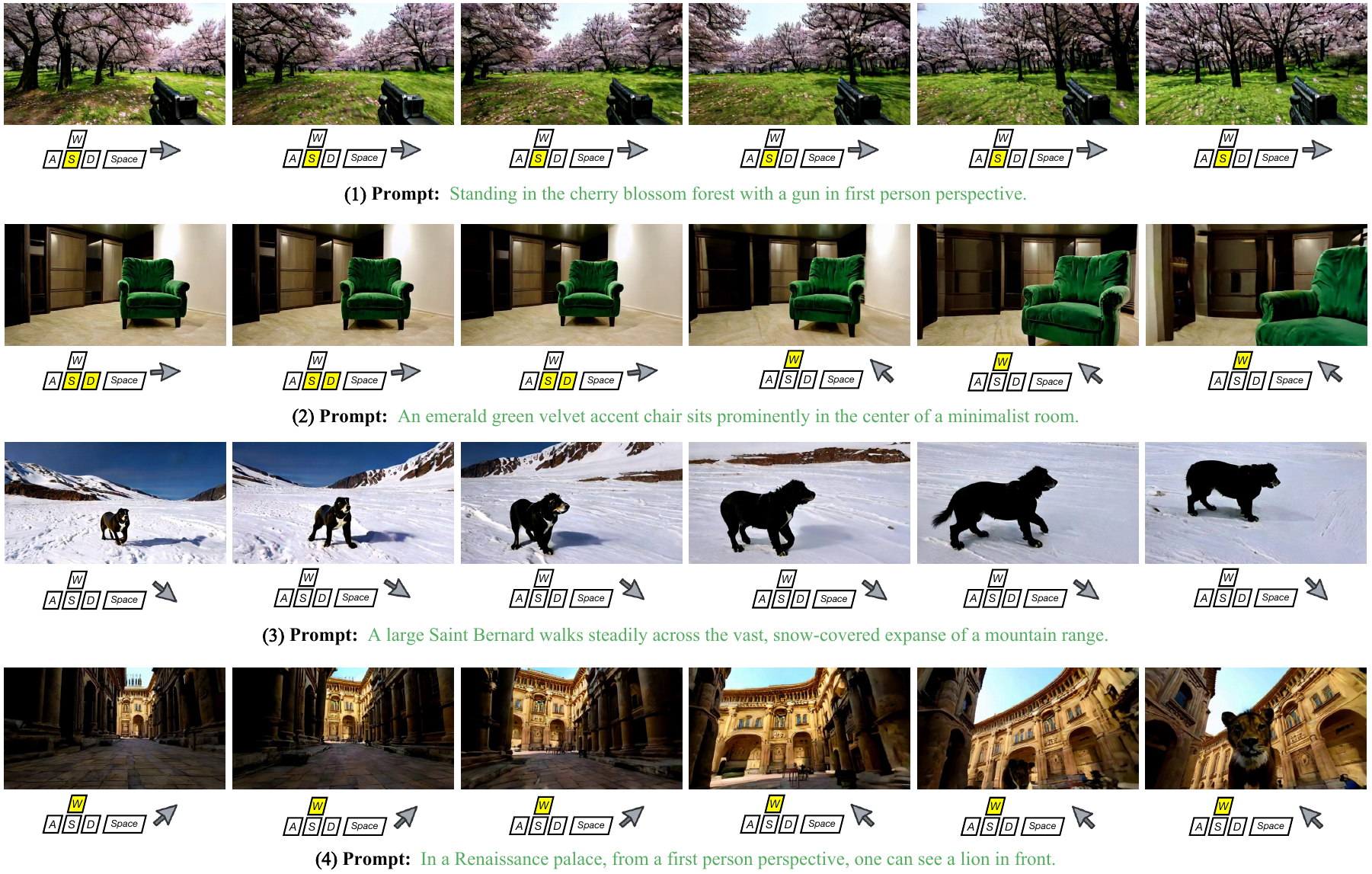}
\vspace{-0.5cm}
\captionsetup{type=figure}
\caption{We propose GameFactory, a framework that leverages the powerful generative capabilities of pre-trained video models for the creation of new games. By learning action control from \textit{a small-scale first-person Minecraft dataset}, this framework can transfer these control abilities to open-domain videos, ultimately allowing the creation of new games within open-domain scenes. As illustrated in (1)-(4), GameFactory supports action control across diverse, newly generated scenes in open domains, paving the way for the development of entirely new game experiences. The yellow buttons indicate \textbf{pressed keys}, and the arrows represent the direction of \textbf{mouse movements}.
}
\label{fig:overview}

\vspace{0.2cm}
}]
\renewcommand*{\thefootnote}{$\ddagger$}
\footnotetext[1]{Corresponding authors. $^*$Equal contribution. $^\dagger$Part of the work done during internship at KwaiVGI, Kuaishou Technology.}

\begin{abstract}
Generative videos have the potential to revolutionize game development by autonomously creating new content. 
In this paper, we present GameFactory, a framework for action-controlled scene-generalizable game video generation. 
We first address the fundamental challenge of action controllability by introducing GF-Minecraft, an action-annotated game video dataset without human bias, and developing an action control module that enables precise control over both keyboard and mouse inputs. We further extend to support autoregressive generation for unlimited-length interactive videos.
More importantly, GameFactory tackles the critical challenge of scene-generalizable action control, which most existing methods fail to address. To enable the creation of entirely new and diverse games beyond fixed styles and scenes, we leverage the open-domain generative priors from pre-trained video diffusion models. To bridge the domain gap between open-domain priors and small-scale game datasets, we propose a multi-phase training strategy with a domain adapter that decouples game style learning from action control. This decoupling ensures that action control learning is no longer bound to specific game styles, thereby achieving scene-generalizable action control.
Experimental results demonstrate that GameFactory effectively generates open-domain action-controllable game videos, representing a significant step forward in AI-driven game generation. 
% Our dataset and code will be publicly available.
\end{abstract}    
\section{Introduction}
\label{sec:intro}
% background: powerfull video diffusion models
Video diffusion models have demonstrated impressive capabilities in video generation~\cite{sora, open-sora-plan, opensora, cogvideox}. These models have the potential to become promising candidates for generative game engines~\cite{gamengine, diamond, oasis, genie, genie2}, which could transform game development by not only significantly reducing manual work in the traditional game industry through automatic content creation, but also enabling the generation of unlimited game content for players to explore. Given this promising direction, it is important to explore how to build a qualified generative game engine.

% motivation: the importance of generalization and existing works fail to cover this topic
Generative game engines are typically implemented as video generation models with action controllability, enabling responses to user inputs like keyboard and mouse interactions~\cite{gamengine, oasis, genie2}. Current research works~\cite{gamengine, diamond, oasis, gamegenx, playgen} mainly focus on specific games such as DOOM~\cite{gamengine,playgen}, Atari~\cite{diamond}, CS:GO~\cite{diamond}, Super Mario Bros~\cite{playgen}, and Minecraft~\cite{oasis}, or use limited game-specific datasets~\cite{gamegenx}. This game-specific approach lacks scene generalization capability, preventing models from creating content beyond existing games and limiting their potential for developing new games. Thus, scene generalization remains a key challenge in advancing generative game engines.

 \begin{figure}[!tbp]
  \centering
  % \vspace{-0.1cm}
  \includegraphics[width=1\linewidth]{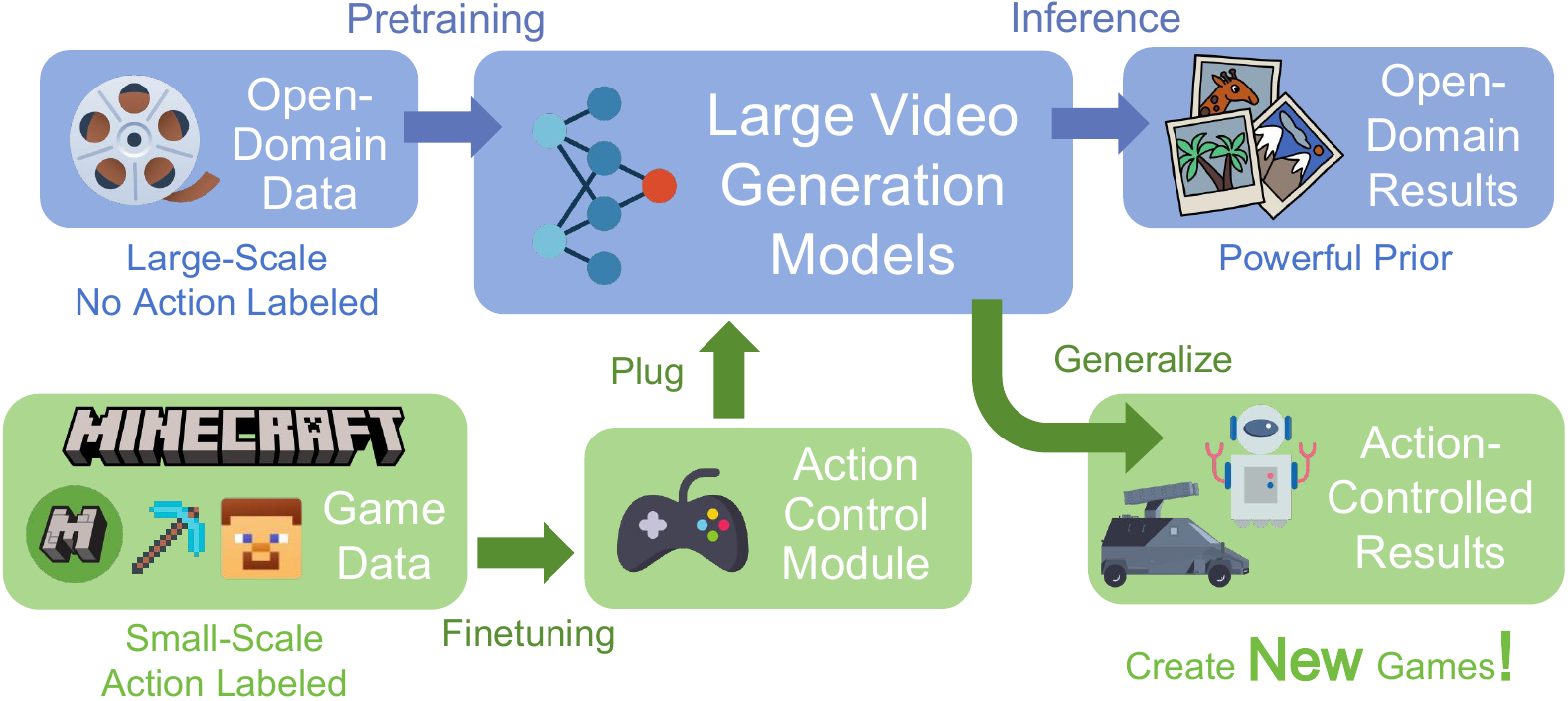}
  % \vspace{-0.7cm}
  \caption{A schematic of our GameFactory creating new games based on pre-trained large video generation models. The \textcolor[rgb]{0.5686, 0.6706, 0.8745}{upper blue section} shows the generative capabilities of the pre-trained model in an open-domain, while the \textcolor[rgb]{0.6706, 0.8392, 0.5569}{lower green section} demonstrates how the action control module, learned from a small amount of game data, can be plugged in to create new games.}
  \vspace{-0.5cm}
\label{fig:concept} 
\end{figure}

% \begin{figure*}[!tbp]
%   \centering
%   % \vspace{-0.1cm}
%   \includegraphics[width=1.5\columnwidth]{figure/concept.pdf}
%   % \vspace{-0.7cm}
%   \caption{A schematic of creating new games based on large pre-trained video generation models.}
%   % \vspace{-0.1cm}
%   \label{fig:concept} 
% \end{figure*}
% core idea: utilizing the pretrained video generation model and minecraft platform
To achieve scene generalization in game videos, collecting large-scale action-annotated video datasets would be the most direct approach. With sufficient data covering all possible scenarios, it could enable arbitrary game scene generation. However, such action annotations are prohibitively expensive and impractical for open-domain videos.
In contrast, open-domain videos are abundant on the Internet, and video generation models trained on these videos contain rich generative priors.
Leveraging these scene-generation priors presents a more feasible path to scene generalization (\textcolor[rgb]{0.5686, 0.6706, 0.8745}{upper blue section} in Fig.~\ref{fig:concept}).
This observation motivates us to ask: With the video generation models pre-trained on large-scale open-domain videos, along with small-scale action-annotated game videos, can we empower scene-generalizable video generation with action control? This is a nontrivial problem, because directly fine-tuning the pre-trained video generation models with action-annotated game videos will lead to degraded scene generalizability (\ie, the model will collapse to the specific domain of the game videos). 

% With the generative video priors from pre-trained models, we only need to learn action-specific knowledge, which can be achieved with limited annotated data and then generalized to open-domain scenarios (\textcolor[rgb]{0.6706, 0.8392, 0.5569}{lower green section} in Fig.~\ref{fig:concept}). 
% Specifically, using a video generation model pre-trained on \textbf{large-scale open-domain video data}, we can train an effective action control module with just \textbf{a small amount of action-annotated data}, achieving controllability while maintaining the powerful generative prior.
% We chose Minecraft~\cite{minedojo} as our action data source for its unique advantages: diverse and complex action space with customizable inputs, frame-by-frame action annotation support, and cost-effective generation of high-quality action data without human bias.

% In this work, we present \textbf{GameFactory}, a framework for \textit{action-controlled and scene-generalizable} game video generation. To achieve action-controlled video generation, we firstly collect a xxx dataset. [elaborate more on the dataset.] Secondly, we design action control mechanisms specifically for continuous mouse movements and discrete keyboard inputs, respectively. The action control modules are xxx [briefly mention technical design] and can be seamlessly injected into the pre-trained video diffusion models without affecting the video generation abilities of the pre-trained models. Lastly, we extend our model to autoregressive generate long videos with xxx.

In this work, we present \textbf{GameFactory}, a framework for \textit{action-controlled and scene-generalizable} game video generation. To achieve action-controlled video generation, we firstly collect an action-annotated game video dataset GF-Minecraft. This dataset, sourced from the Minecraft platform~\cite{minedojo}, features an unbiased action distribution, diverse scenes, and text annotations (Sec.~\ref{subsec:data_collect}). Secondly, we design action control mechanisms for continuous mouse movements and discrete keyboard inputs, respectively. The action control modules can be seamlessly injected into the pre-trained video diffusion models without affecting the video generation abilities of the pre-trained models (Sec.~\ref{subsec:module}). Lastly, we extend our model to autoregressive long video generation by allowing varying noise levels across different frames, continuously conditioning on previously generated frames to generate new video frames (Sec.~\ref{subsec:ar}).

However, directly fine-tuning the pretrained model on Minecraft data not only imparts action control capabilities, but also risks embedding a distinct Minecraft style into the generated games, thereby compromising the model's open-domain generalization. To address this challenge, our key idea is to disentangle the learning of game style and action control with different modules and parameters, so that we can remove the game style without affecting the action controllability for generating open-domain videos with action control (Sec.~\ref{subsec:decoupled_model}).
Specifically, we propose a domain adapter to fit Minecraft style, combined with a multiphase decoupled training strategy (Sec.~\ref{subsec:training_strategy}).

In summary, our key contributions include: 
\begin{itemize} 
    % \item We propose GameFactory for open-domain scene generalization in game video generation \xihui{action-controlled and scene-generalizable video generation}. To overcome game-specific limitations, we leverage pre-trained video diffusion models with a multiphase training strategy that decouples game style from action control, enabling diverse interactive game environments beyond existing games.
    % \item We introduce GF-Minecraft, a new action-annotated dataset for generative game engines. To address the challenge of human bias in action annotation, we curate a high-quality and diverse dataset from Minecraft.
    % \item We develop an autoregressive generation mechanism that enables unlimited-length action-controlled game videos, meeting the critical requirements of continuous gameplay in practical game applications.

    \item We propose GameFactory for action-controlled and scene-generalizable video generation, enabling diverse interactive game generation beyond existing games.
    \item We introduce GF-Minecraft, an action-annotated game video dataset featuring unbiased action distributions, diverse scenes, and text descriptions. We also design dedicated mechanisms for action control and autoregressive long video generation.
    \item We propose a domain adapter with multiphase decoupled training strategy that disentangles game style from action control, enabling open-domain video generation while preserving action controllability.
\end{itemize}

\section{Related Work}
\subsection{Video Diffusion Models}
With the rise of diffusion models~\cite{ddpm, sde, score}, significant progress has been made in visual generation, spanning both image~\cite{sd, pixart, sdxl} and video generation~\cite{videocrafter2, svd, opensora, open-sora-plan, cogvideox}. In particular, recent advancements in video generation have seen a shift from the U-Net architecture to Transformer-based architectures, enabling video diffusion models to generate highly realistic, longer-duration videos~\cite{dit, sora}. The quality of these generated videos has led to the belief that video diffusion models are capable of understanding and simulating real-world physical rules~\cite{unisim, sora, video-new-language}, suggesting their potential use as world models in areas such as autonomous driving and robotics.
% Of course, this requires the combination with other techniques, such as conditional control~\cite{motionctrl, cameractrl}, image-to-video mechanisms~\cite{dynamicrafter}, and auto-regressive generation mechanisms~\cite{diffusion-forcing}.

\subsection{Controllable  Video Generation}
Text descriptions alone often provide limited control in text-to-video models, leading to ambiguous outputs. To enhance control, several methods have introduced additional control signals. For instance, approaches such as~\cite{guo2025sparsectrl, ni2023conditional, dynamicrafter} incorporate images as control signals for the video generator, improving both video quality and temporal relationship modeling. Direct-a-Video~\cite{yang2024direct} uses a camera embedder to adjust camera poses, but its reliance on only three camera parameters limits control to basic movements, like left panning. In contrast, MotionCtrl~\cite{motionctrl} and CameraCtrl~\cite{cameractrl} offer more complex and nuanced control over camera poses in generated videos.

\subsection{Game Video Generation}
Early works~\cite{drivegan, gamegan, caddy, pe, pgm} began exploring game generation using generative models like GAN, but were primarily limited by the generative capability of these models.
Recently, given the promising capabilities of video generation models, researchers have begun exploring their application in game generation~\cite{genie, diamond, gamengine, gamegenx, oasis, matrix, playgen, genie2, wham}.
 Genie~\cite{genie} proposes a foundation model for playable world based on video generation.
 DIAMOND~\cite{diamond}, GameNGen~\cite{gamengine}, Oasis~\cite{oasis} and PlayGen~\cite{playgen} leverage diffusion-based world modeling for specific games like Atari~\cite{diamond}, CS:GO~\cite{diamond}, DOOM~\cite{gamengine, playgen}, Minecraft~\cite{oasis} and Super Mario Bro~\cite{playgen}.
 GameGenX~\cite{gamegenx} introduces OGameData for game video generation and control.
 However, these works suffer from overfitting to specific games or datasets, exhibiting limited scene generalization capabilities.
% Other works, such as Genie 2~\cite{genie2} and Matrix~\cite{matrix}, have discussed the issue of achieving control generalization. However, Genie 2 still relies on collecting large amounts of high-cost action-labeled data to learn sufficient knowledge about game scenarios, while Matrix has limited testing scenes and only evaluates generalization on three basic actions (left turn, right turn, and acceleration) in racing games.
Recent works like Genie 2~\cite{genie2} and Matrix~\cite{matrix} have made valuable progress towards control generalization in game video generation. While Genie 2 demonstrates impressive results through extensive action-labeled data collection, and Matrix shows promising generalization capabilities in racing games, there remains room for improvement in achieving broader scene generalization with more diverse action controls. 
 
In contrast, our proposed GameFactory addresses scene generalization by utilizing pretrained video model priors and the easily-accessible and low-cost game dataset GF-Minecraft, with experiments validating its effectiveness across diverse open-world scenarios and a much more complex action space (including forward/backward/left/right movement, jumping, and acceleration/deceleration).

\section{Preliminaries}
\label{sec:preliminary}
We adopt a transformer-based latent video diffusion model~\cite{dit, sora} as the backbone. Let $\mathbf{X}$ represent a video sequence. To reduce modeling complexity, an Encoder $E(\cdot)$ compresses the video spatially and temporally into a latent representation $\mathbf{Z}=E(\mathbf{X})$. With temporal compression ratio $r$, a $(1+rn)$-frame video is compressed into $(1+n)$ latent frames. Denoting the $i$-th frame as $\mathbf{x}^{i}$ and $i$-th latent as $\mathbf{z}^{i}$, we have $\mathbf{X}=[\mathbf{x}^{0}, \mathbf{x}^{1}, ..., \mathbf{x}^{rn}]$ and $\mathbf{Z}=[\mathbf{z}^{0}, \mathbf{z}^{1}, ..., \mathbf{z}^{n}]$. 
During training, noise is added to the clean latent $\mathbf{Z}_0$ to get noisy latent $\mathbf{Z}_t$ at timestep $t$. 
$\boldsymbol{\epsilon}$ is the added random noise.
% The noise predictor is trained using the following loss function:
% \begin{equation}
%     \mathcal{L}(\boldsymbol{\theta})=\mathbb{E}[||\boldsymbol{\epsilon}_{\boldsymbol{\theta}}(\mathbf{Z}_t,\mathbf{p},t)-\boldsymbol{\epsilon}||_2^2],
% \end{equation}
% where $\boldsymbol{\theta}$ denotes the model parameters and $\mathbf{p}$ is the prompt input. During inference, with the trained noise predictor, we can sample clean latent $\mathbf{Z}_0$ from a noisy latent $\mathbf{Z}_T$. The predicted latent $\mathbf{Z}_0$ is then decoded back to video $\mathbf{X}$ through $D(\cdot)$: $\mathbf{X}=D(\mathbf{Z}_0)$.
When considering action control, the action $\mathbf{A}=[\mathbf{a}^1, \mathbf{a}^2, ..., \mathbf{a}^{rn}]$, where $\mathbf{a}^i$ represents the action taken at timestep $(i-1)$ to transfer from $\mathbf{x}^{i-1}$ to $\mathbf{x}^{i}$. The corresponding action-conditioned loss function is:
\begin{equation}
    \mathcal{L}_{\mathbf{a}}(\boldsymbol{\phi})=\mathbb{E}[||\boldsymbol{\epsilon}_{\boldsymbol{\phi}}(\mathbf{Z}_t,\mathbf{p}, \mathbf{A},t)-\boldsymbol{\epsilon}||_2^2],
    \label{eq:conditional_training_loss}
\end{equation}
where $\boldsymbol{\phi}$ means the model parameters, and $\mathbf{p}$ is the prompt input. During inference, we can sample clean latent $\mathbf{Z}_0$ from a noisy latent $\mathbf{Z}_T$. The predicted latent $\mathbf{Z}_0$ is then decoded back to video $\mathbf{X}$ through $D(\cdot)$: $\mathbf{X}=D(\mathbf{Z}_0)$.

\section{Action-Controlled Video Generation}
This section introduces implementing an action-controlled video generation model, which is the foundation for Sec.~\ref{sec:generalization}. Sec.~\ref{subsec:data_collect} describes our collected game video dataset and its advantages; Sec.~\ref{subsec:module} introduces how to implement the action control module for responding to player actions; Sec.~\ref{subsec:ar} presents the method for autoregressive long video generation, which is the key to creating playable game.

\subsection{GF-Minecraft Dataset}
\label{subsec:data_collect}
% \input{fig_tab_subtex/action_control_module}
% \subsection{Action-Controllable Video Collection}

For our action-controllable video generation model to simulate real game engines, the training data should satisfy three key requirements: (1) easily accessible with customizable action inputs to enable cost-effective data collection; (2) action sequences free from human bias, allowing extreme and low-probability action combinations to support arbitrary action inputs; (3) diverse game scenes with corresponding textual descriptions to learn scene-specific physical dynamics. Existing datasets like VPT~\cite{vpt} are collected from human gameplay videos, which inherently contain human behavioral biases and lack text descriptions of the scenes. To address these limitations, we introduce our GF-Minecraft dataset, where `GF' stands for our method GameFactory and `Minecraft' refers to the game name. 
The advantages of our dataset are summarized as follows and its details can be found in the supplementary materials:

\vspace{2pt}\noindent\textbf{\ding{113} Minecraft as an Accessible Data Source.} 
We leverage Minecraft as our data collection platform for its comprehensive API that captures detailed environmental snapshots, enabling large-scale data collection with action annotations. The game also offers extensive scenes, navigable areas, diverse action space, and an open-world environment. By executing predefined action sequences, we collected 70 hours of gameplay video as GF-Minecraft Dataset.

\vspace{2pt}\noindent\textbf{\ding{113} Collecting Videos with Unbiased Action.}
Existing Minecraft datasets, such as VPT~\cite{vpt}, are collected from real human gameplay, resulting in biased action distributions that favor common human behaviors. Models trained on these datasets overlook rare action combinations, such as moving backward, jumping in place, or standing still with mouse movements. To eliminate such biases, we decompose keyboard and mouse inputs into atomic actions and ensure their balanced distribution. We also randomize the frame duration of each atomic action to avoid temporal bias.

\vspace{2pt}\noindent\textbf{\ding{113} Diverse Scenes with Textual Descriptions.}
To enhance dataset diversity, we captured videos across different scenes, weather conditions, and times of day. We segmented the videos and annotated them with textual descriptions using an efficient multimodal LLM, MiniCPM~\cite{minicpm}.

\subsection{Action Control Module}
\label{subsec:module}
\begin{figure}[t]
  \centering
%   \vspace{-0.1cm}
  \includegraphics[width=1\linewidth]{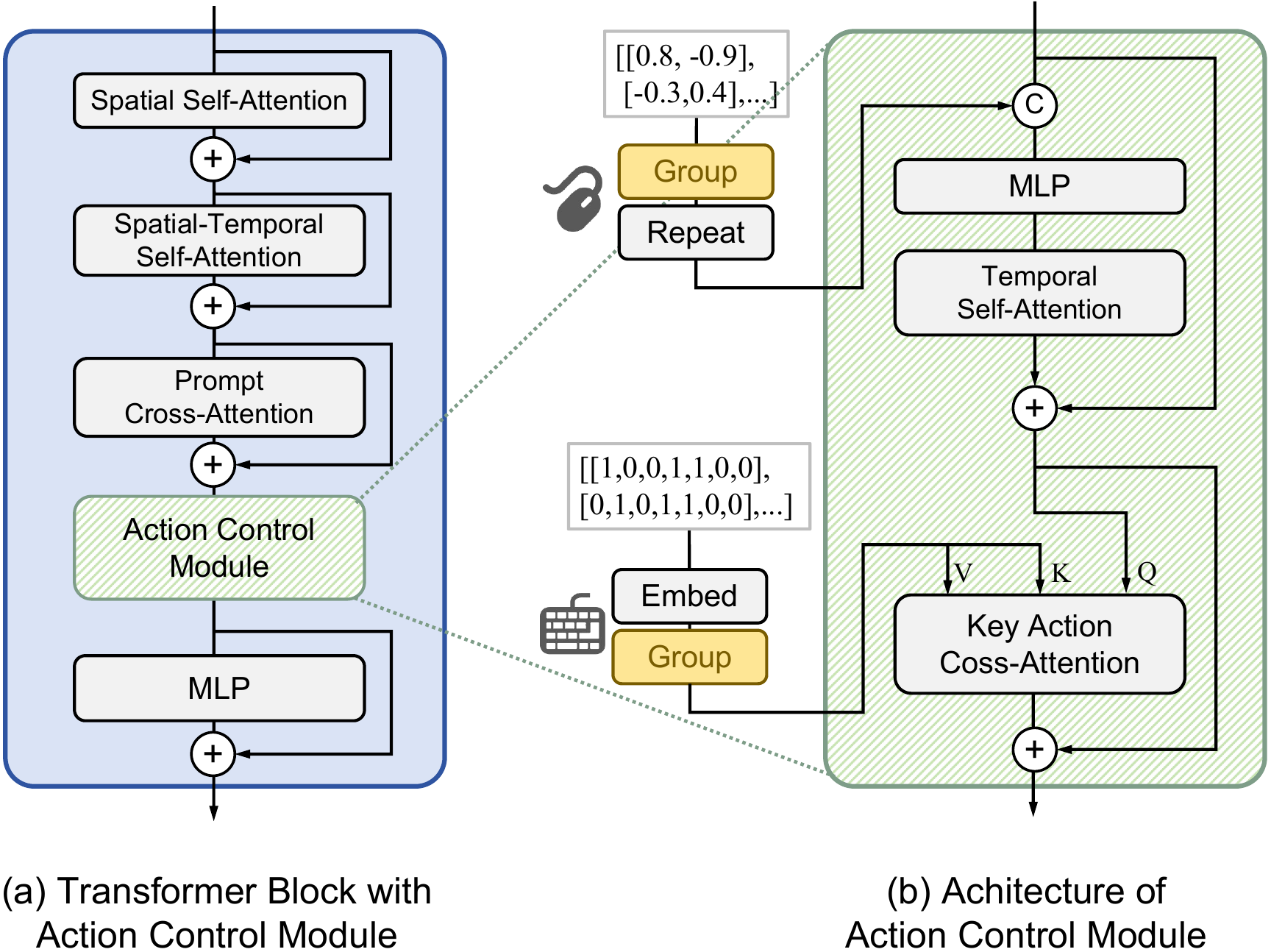}
  % \vspace{-0.6cm}
  \caption{
(a) Integration of Action Control Module into transformer blocks of the video diffusion model. 
(b) Different control mechanisms for continuous mouse and discrete keyboard inputs. 
  }
  \vspace{-0.5cm}
\label{fig:action_arch1} 
\end{figure}
\begin{figure}[t]
  \centering
%   \vspace{-0.1cm}
  \includegraphics[width=1\linewidth]{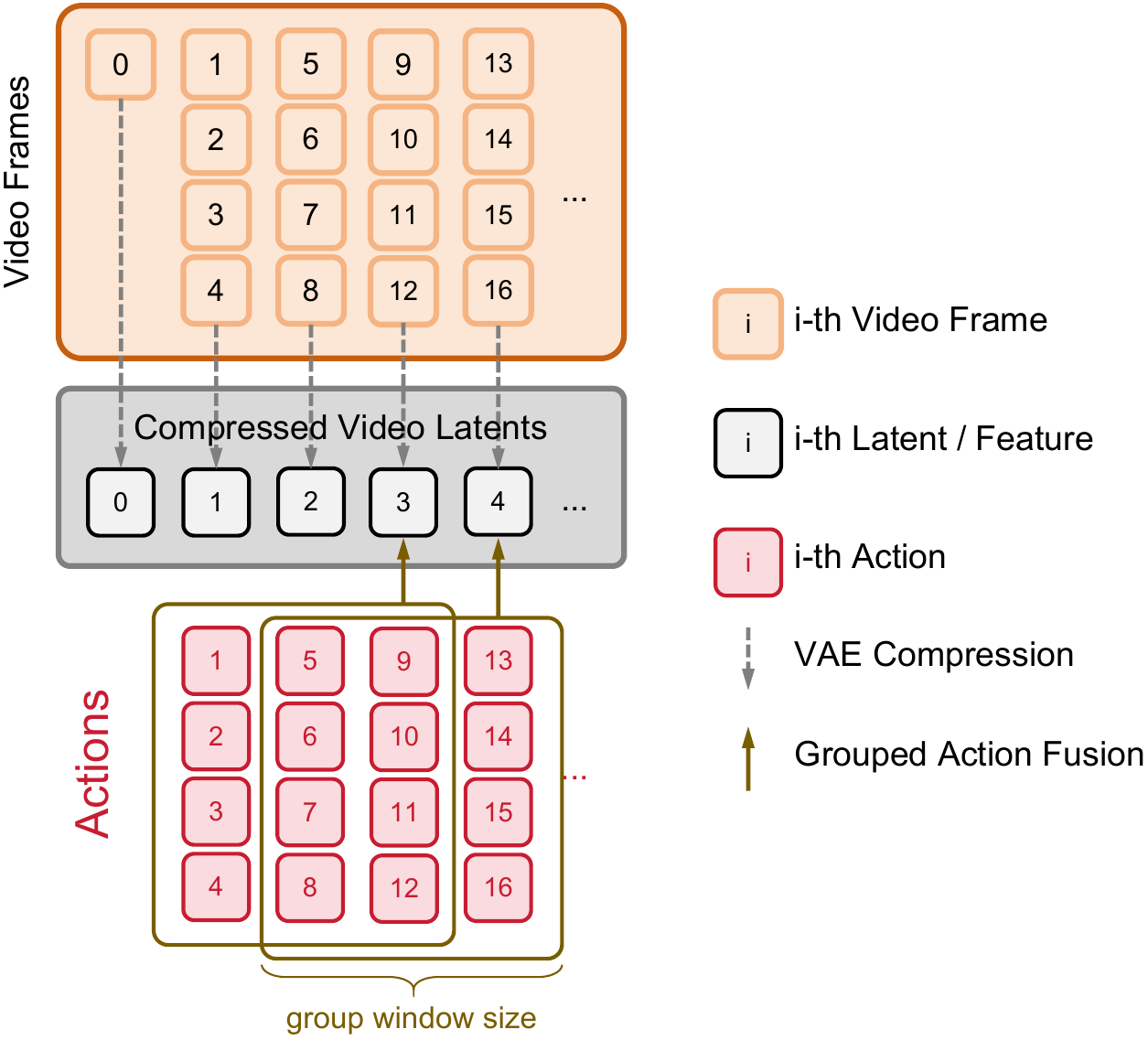}
  % \vspace{-0.6cm}
  \caption{
Due to temporal compression (compression ratio $r=4$), the number of latent features differs from the number of actions, causing granularity mismatch during fusion. Grouping aligns these sequences for fusion. Additionally, the $i$-th latent feature can fuse with action groups within a previous window (window size $w=3$), accounting for delayed action effects (e.g., `jump' key affects several subsequent frames).
  }
  \vspace{-0.5cm}
\label{fig:action_arch2} 
\end{figure}
% For the action control module, we will introduce the situation when adding an action control module to a Transformer block. 
We incorporate action control modules into the transformer blocks of the video diffusion model to enable action-controllable generation.
The architecture of the transformer block is shown in Fig.~\ref{fig:action_arch1} (a), and the structure of the action control module is shown in Fig.~\ref{fig:action_arch1} (b). 
Suppose that the input action includes both continuous mouse movement action $\mathbf{M}\in\mathbb{R}^{rn\times d_1}$, and discrete keyboard action $\mathbf{K}\in\mathbb{R}^{rn\times d_2}$. The intermediate feature in the Transformer is denoted as $\mathbf{F}\in\mathbb{R}^{(n+1)\times l \times c}$.
% \begin{align}
%     \mathbf{M}&=[\mathbf{m}^{1}, \mathbf{m}^{2}, ..., \mathbf{m}^{rn}]\in\mathbb{R}^{rn\times d_1},\\
%     \mathbf{K}&=[\mathbf{k}^{1}, \mathbf{k}^{2}, ..., \mathbf{k}^{rn}]\in\mathbb{R}^{rn\times d_2},\\
%     \mathbf{F}&=[\mathbf{f}^{0}, \mathbf{f}^{1}, ..., \mathbf{f}^{n}]\in\mathbb{R}^{(n+1)\times l \times c},
% \end{align}
% where 
% $\mathbf{m}^{i}$ represents the $i$-th mouse action, $\mathbf{k}^{i}$ represents the $i$-th keyboard action, $\mathbf{f}^{i}$ represents the feature of the $i$-th latent frame. 
The total number of video frames is $(1+rn)$, compressed to $(1+n)$ latent frames, where $r$ denotes the temporal compression ratio. $d_1$ and $d_2$ denote the dimension numbers of actions, $l$ is the length of token sequence and $c$ is the number of feature channels.

\vspace{2pt}\noindent\textbf{\ding{113} Grouping Actions with a Sliding Window.} 
Due to the temporal compression ratio $r$, the number of actions ($rn$) differs from the number of features ($n+1$), creating a granularity mismatch for action-feature fusion. As shown in Fig.~\ref{fig:action_arch2}, we address this by grouping actions using a sliding window of size $w$. For the $i$-th feature $\mathbf{f}^{i}$, we consider actions within $[\mathbf{a}^{r\times(i-w+1)}, ..., \mathbf{a}^{ri}]$. This window design captures delayed action effects, such as how a jump command influences multiple subsequent frames. For out-of-range indices, boundary actions are used as padding.
For mouse movement $\mathbf{M}$, the grouped action is $\mathbf{M}_{group}\in\mathbb{R}^{(n+1)\times rw \times d_1}$.
As for keyboard actions $\mathbf{K}$, we first learn the embedding of actions and add the positional encoding.
After that, we perform a grouping operation on the action embeddings to get $\mathbf{K}_{group}\in\mathbb{R}^{(n+1)\times rw\times c}$.

\vspace{2pt}\noindent\textbf{\ding{113} Mouse Movements Control.}
To fuse the grouped mouse action $\mathbf{M}_{group}$ with feature $\mathbf{F}$, we first reshape it from $\mathbb{R}^{(n+1)\times rw \times d_1}$ to $\mathbb{R}^{(n+1)\times 1\times rwd_1}$. Then we repeat it in the dimension of the token sequence length to get $\mathbf{M}_{repeat}\in\mathbb{R}^{(n+1)\times l\times rwd_1}$,
and concatenate it with $\mathbf{F}$ along with the channel dimension to get $\mathbf{F}_{fused}\in \mathbb{R}^{(n+1)\times l\times (c+rwd_1)}$.
After that, further learning on $\mathbf{F}_{fused}$ is conducted through a layer of MLP and a layer of temporal self-attention.

\vspace{2pt}\noindent\textbf{\ding{113} Keyboard Actions Control.}
For discrete keyboard control, we perform a cross-attention calculation between the grouped action embeddings $\mathbf{K}_{group} \in \mathbb{R}^{(n+1) \times rw \times c}$ and $\mathbf{F} \in \mathbb{R}^{(n+1) \times l \times c}$, similar to the prompt cross-attention between text and $\mathbf{F}$. Specifically, $\mathbf{K}_{group}$ serves as the key and value in the attention, while $\mathbf{F}$ functions as the query.

% \vspace{2pt}\noindent\textbf{\ding{113} Action-Specific Control Module.}
% Our experiments indicate concatenation works better for continuous actions (e.g., mouse movements) while cross-attention suits discrete actions (e.g., keyboard inputs). For continuous actions, concatenation preserves the magnitude information critical for control, whereas cross-attention's similarity computations and normalization would diminish these numerical differences. For discrete actions, cross-attention's effectiveness aligns with its successful applications in text prompt control, demonstrating its strength in categorical processing.

\subsection{Autoregressive Generation of Long Action-Controllable Game Videos}
\label{subsec:ar}
 \begin{figure}[!tbp]
  \centering
  % \vspace{-0.1cm}
  \includegraphics[width=1\linewidth]{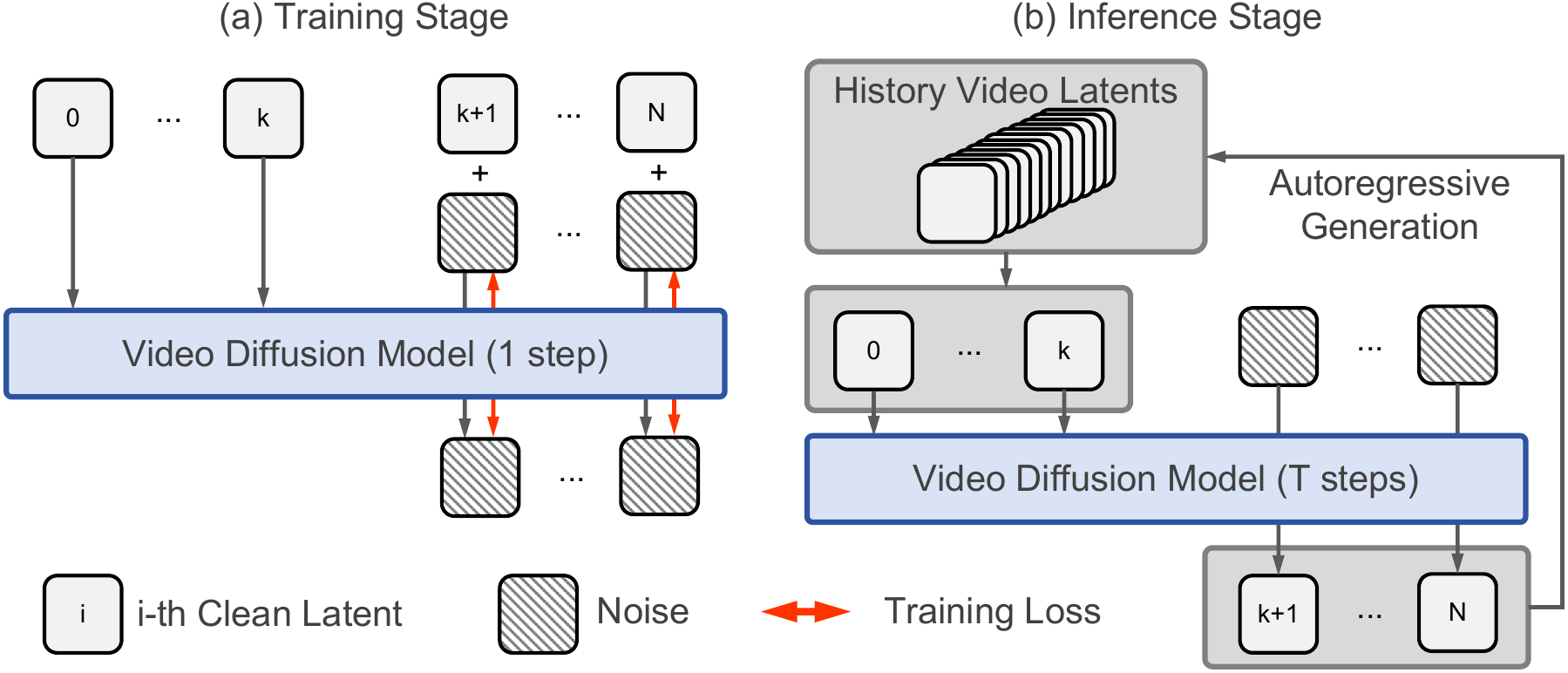}
  % \vspace{-0.7cm}
  \caption{
  Illustration of autoregressive video generation. The frames from index $0$ to $k$ serve as conditional frames, while the remaining $N-k$ frames are for prediction, with $k$ randomly selected. (a) Training stage: Loss computation and optimization focus only on the noise of predicted frames. (b) Inference stage: The model iteratively selects the latest $k+1$ frames as conditions to generate $N-k$ new frames, enabling autoregressive generation.
  }
  \vspace{-0.5cm}
\label{fig:ar_gen} 
\end{figure}
\begin{figure*}[ht]
  \centering
%   \vspace{-0.1cm}
  \includegraphics[width=1\linewidth]{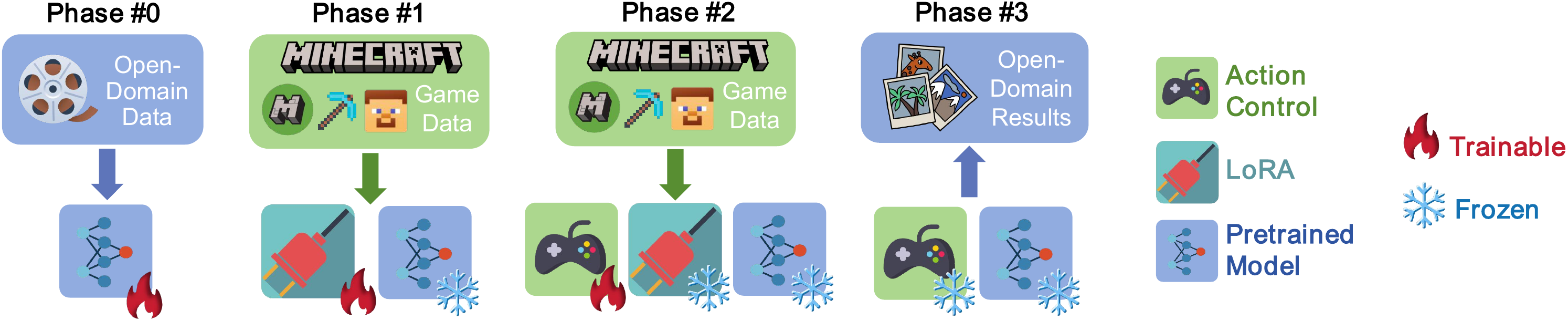}
  % \vspace{-0.6cm}
  \caption{
%   The overall pipeline of our proposed multi-phase training strategy for scene generalization.
% \textbf{Phase \#0}: pretraining a pretrained large video generation model on open-domain data.
% \textbf{Phase \#1}: finetuning with LoRA for game video data.
% \textbf{Phase \#2}: training the action control module while fixing other parameters.
% \textbf{Phase \#3}: inference to get generalized open-domain results. 
% In order to decouple the learning of game-specific style from the learning of action control, \textbf{Phase \#1} focuses on learning style information from a specific game dataset, while \textbf{Phase \#2} specializes in learning action control independently of any particular game style. This approach preserves the open-domain generative capabilities learned in \textbf{Phase \#0}, enabling scene generalization in \textbf{Phase \#3}.
\textbf{Phase \#0}: pretraining a video generation model on open-domain data.
\textbf{Phase \#1}: finetuning with LoRA for game video data.
\textbf{Phase \#2}: training the action control module while fixing other parameters.
\textbf{Phase \#3}: inference for action-controlled open-domain generation.
To decouple style learning from action control, \textbf{Phase \#1} learns game-specific style while \textbf{Phase \#2} focuses on style-independent action control. This design preserves the open-domain capabilities from \textbf{Phase \#0}, enabling generalization in \textbf{Phase \#3}.
  }
  \vspace{-0.2cm}
\label{fig:training_strategy} 
\end{figure*}

Current video generation methods introduced in Sec.~\ref{sec:preliminary} are limited to fixed-length outputs, which is inadequate for practical game applications requiring continuous video streams. To address this, we develop an autoregressive approach that generates multiple frames per step based on previous outputs, enabling efficient long video generation.

% One established approach for autoregressive video generation follows next-token prediction~\cite{videopoet,magvitv2,videogpt}, but lacks accessible strong model parameters. 
While video diffusion transformer~\cite{sora,dit,open-sora-plan,opensora} offers superior generation quality, it typically operates in a full-sequence manner. 
Inspired by Diffusion Forcing~\cite{diffusion-forcing}, we modify the video diffusion transformer to enable autoregressive generation. Specifically, unlike standard diffusion models that require identical noise levels across frames, our approach allows different noise levels where later frames can have more noise while depending on earlier frames with less noise. This varying noise schedule ensures that earlier frames are generated first, allowing subsequent frames to be conditioned on them in an autoregressive manner.

As shown in Fig.~\ref{fig:ar_gen} (a), during training with $N+1$ frame latents (from $0$ to $N$), we randomly select the first $k+1$ frames as conditions without adding noise, while adding noise only to the remaining $N-k$ frames for noise prediction training. Although the first $k+1$ frames are input to the model, since they are assumed to be already generated, their predicted noise outputs are not utilized. To improve training efficiency, we focus on computing training losses only for the $N-k$ frames that require prediction.
As for inference, as shown in Fig.~\ref{fig:ar_gen} (b), after full-sequence generation of the first $N+1$ frame latents, we can autoregressively generate new $N-k$ frames by selecting the most recent $k+1$ frames from history video latents as conditions for each subsequent generation, and merge them into the history video latents. This process can be repeated to achieve infinite-length video generation.
Unlike conventional next-frame generation methods~\cite{diffusion-forcing,gamengine,oasis}, our approach supports \textbf{multi-frame generation in one step}, greatly reducing the time for long video generation.

% \vspace{2pt}\noindent\textbf{\ding{113} Implementation Details.} 
% Since the model structure remains unchanged and only the noise levels vary across frames, the training configuration remains identical to full-sequence generation, including action control. We also add small noise perturbations (equivalent to step $15$/$1000$) to conditional frames, which helps reduce error accumulation in long-term generation~\cite{gamengine, oasis}.
% The long video results for action control can be found on our project page.

\section{Open-Domain Game Scene Generalization}
\label{sec:generalization}
This section introduces how to generalize learned action control capabilities to open-domain scenes. As analyzed in Sec.~\ref{sec:intro}, while pre-trained video models offer rich generative priors for open-domain generation, directly fine-tuning them with game data leads to undesired style bias: the outputs inherit the visual style of training data while learning action control. To address this, we propose a Style-Action Decoupled Model with Domain Adapter (Sec.~\ref{subsec:decoupled_model}) and a multi-phase training strategy (Sec.~\ref{subsec:training_strategy}).

\subsection{Style-Action Decoupling with Domain Adapter}
\label{subsec:decoupled_model}

To prevent action control capabilities from being bound to specific game styles, our key insight is to disentangle the learning of game style and action control through different modules and parameters. Specifically, while using the proposed action control module to learn action controllability, we employ an independent domain adapter to capture game-specific visual styles. The domain adapter is implemented using LoRA~\cite{lora}, which efficiently learns specific styles and can be plugged in and out without affecting the open-domain generation priors of the original model. To effectively train these decoupled components, we need a carefully designed training strategy that ensures the style adaptation and action control learning remain independent by learning them in different training phases. We detail this multi-phase training approach in the next section.

\subsection{Multi-Phase Training Strategy}
\label{subsec:training_strategy}
\begin{table*}[!ht]\footnotesize
    \centering
    \renewcommand{\arraystretch}{1.5} % 调整行高
    \renewcommand{\cellalign}{vh} % 单元格内容垂直和水平居中
    \newcommand{\gou}{\textcolor{ForestGreen}{\ding{52}}}
    \newcommand{\cha}{\textcolor{Red}{\ding{55}}}
    \begin{tabular}{>{\centering\arraybackslash}p{2.3cm}|
    >{\centering\arraybackslash}p{1.7cm}|
    >{\centering\arraybackslash}p{1.7cm}|
    >{\centering\arraybackslash}p{1.7cm}|
    >{\centering\arraybackslash}p{1.7cm}|
    >{\centering\arraybackslash}p{1.7cm}|
    >{\centering\arraybackslash}p{1.7cm}|
    >{\centering\arraybackslash}p{1.7cm}
}
        \toprule
        & \textbf{DIAMOND}~\cite{diamond} &\textbf{GameNGen}~\cite{gamengine} & \textbf{GameGenX}~\cite{gamegenx} & \textbf{Oasis}~\cite{oasis} & \textbf{Matrix}~\cite{matrix} & \textbf{Genie 2}~\cite{genie2} & \textbf{GameFactory} \\
        \hline\hline
        Release Time & NeurIPS 2024 & ICLR 2025 & ICLR 2025 & 2024.10.31 & 2024.12.4 & 2024.12.4 & - \\
        \hline
        Game Sources & Atari, CS:GO & DOOM & AAA Games& Minecraft & AAA Games & Unknown & Minecraft \\
        \hline
        Resolution & $280\times150$ & $240$p & $720$p & $640\times360$ & $720$p & $720$p & $640\times360$\\
        \hline
        Control Granularity & Frame-level & Frame-level & Video-level & Frame-level & Frame-level & Frame-level & Frame-level\\
        \hline
        Technical Paper & \gou & \gou & \gou & \cha & \gou & \cha & \gou \\
        \hline
        Testable Model & \gou & \cha & \cha & \cha & \cha & \cha & - \\
        \hline
        Available Dataset & \cha & \cha & \gou & \cha & \cha & \cha & \gou \\
        \hline
        Action Space & 18 Keys & Key & Instruction & Key + Mouse & 4 Keys & Key+Mouse & 7 Keys+Mouse \\
        \hline
        Scene Generalizable & \cha & \cha & \cha & \cha & \gou & \gou & \gou \\
        \bottomrule
    \end{tabular}
    \caption{Comparison with recent related works. Scene-generalizable action control is the core contribution of GameFactory, being the only technical paper that validates on complex action space and conducts extensive testing across open-domain scenarios.}
    \vspace{-0.5cm}
    \label{tab:compare_works}
\end{table*}

As illustrated in Fig.~\ref{fig:training_strategy}, our training process consists of \textbf{Phase \#0} (model pretraining) and the following phases:

\noindent\textbf{\ding{113} Phase \#1: Tune LoRA to Fit Game Videos.} We fine-tune the pre-trained video diffusion model using LoRA to adapt it to specific game video while preserving most original parameters. This produces a model specialized for the target game domain. Better style adaptation in this phase allows the next phase to focus purely on action control, reducing style-control entanglement.

\noindent\textbf{\ding{113} Phase \#2: Tune Action Control Module.} We freeze both pre-trained parameters and LoRA, only training the action control module with game videos and action signals. Since \textbf{Phase \#1} has handled style adaptation through LoRA, the training loss now focuses less on style learning. This allows the model to concentrate on action control learning, as it becomes the main contributor to minimizing the diffusion loss. Such separation enables style-independent control that can generalize to open-domain scenarios.

\noindent\textbf{\ding{113} Phase \#3: Inference on Open Domain.} During inference, we remove the LoRA weights for game style adaptation, keeping only the action control module parameters. Thanks to the decoupling in previous phases, the action control module can now work independently of specific game styles, enabling controlled game video generation across open-domain scenarios.

\begin{table*}[ht]\footnotesize

\vspace{-0.2cm}
\tabcolsep=0.09cm
\center
\begin{tabular}{c  c | c c c c c | c c c c c | c c c c c}
\toprule
  \multicolumn{2}{c}{Control Module}  & \multicolumn{5}{c}{Only-Key}& \multicolumn{5}{c}{Mouse-Small}& \multicolumn{5}{c}{Mouse-Large}\\
 Key & Mouse & Cam$\downarrow$ & Flow$\downarrow$ & CLIP$\uparrow$ & FID$\downarrow$ & FVD$\downarrow$ & Cam$\downarrow$ & Flow$\downarrow$ & CLIP$\uparrow$ & FID$\downarrow$ & FVD$\downarrow$ & Cam$\downarrow$ & Flow$\downarrow$ & CLIP$\uparrow$ & FID$\downarrow$ & FVD$\downarrow$   \\
\hline
\hline
 Cross-Attn & Cross-Attn & 0.0527 & 8.67 & \textbf{0.3313} & 107.13 & 814.05 & 0.0798 & 20.46 & 0.3137 & \textbf{125.67} & 1203.29 & 0.1362 & 325.18 & 0.3103 & 167.37 & \textbf{1383.92} \\
 Concat & Concat & 0.0853 & 22.37 & 0.3277  & \textbf{103.89}  & \textbf{786.50} & 0.0756 & 19.18 & 0.3159 & 133.42 & 1151.71 & 0.1179 & 258.93 & \textbf{0.3123} & 145.74 & 1405.47 \\
 \textbf{Cross-Attn} & \textbf{Concat} & \textbf{0.0439} & \textbf{7.79} & 0.3292 & 105.28 & 795.03 & \textbf{0.0685} & \textbf{18.64} & \textbf{0.3184} & 127.84 & \textbf{1032.98} & \textbf{0.1021} & \textbf{249.54} & 0.3107 & \textbf{139.91} & 1420.89 \\
\bottomrule

\end{tabular}
\caption{Results of the ablation study on action control mechanisms. The findings indicate that an optimal approach for the action control module is to use cross-attention for discrete action control and concatenation for continuous action control.}
\vspace{-0.5cm}
\label{table:ablation}
\end{table*}

\section{Experiments}
\subsection{Implementation Details}
\vspace{1pt}\noindent\textbf{\ding{113} Pretrained Model Setting.}
Our experiments are based on an internal $1$B-sized transformer-based text-to-video diffusion model for research purpose, which is distilled from a larger pretrained video diffusion model and possesses strong generative priors in the open domain. The resolution of the game videos is $360\times640$.
%, with a frame rate of $15$
%, and the text-to-video generation length is $77$ frames. 
The temporal compression rate of the VAE is $r=4$.

\vspace{1pt}\noindent\textbf{\ding{113} Training and Inference Setting.}
Each phase of fine-tuning or training requires about two-four days of training on $8$ A100 GPUs with a batch size of $64$. The hyper parameters for LoRA finetuning can be referenced as $rank=128$, with a learning rate of 1e-4. The learning rate for training the action control module is 1e-5.
We only apply classifier free guidance~\cite{cfg} to the text prompt conditional input and use DDIM~\cite{ddim} sampling with $50$ sampling steps.

\vspace{1pt}\noindent\textbf{\ding{113} Evaluation Setting.} 
We retained $5\%$ of our collected, segmented dataset as a testset, excluding it from training, and selected three subsets for ablation study: (1) \textbf{only-key}: contains only keyboard actions, designed to test the model’s ability to follow discrete actions; (2) \textbf{mouse-small}: includes small-scale continuous mouse movements; and (3) \textbf{mouse-large}: includes large-scale continuous mouse movements. Additionally, for qualitative experiments, we support custom combinations of input actions, allowing us to test complex or rare action combinations.
We use these evaluation metrics: 
(1) \textbf{Flow}: calculates the optical flow of the generated video to reflect its dynamics, assessing action-following ability by measuring mean square error to the optical flow of the reference video; 
(2) \textbf{Cam}: computes the Euclidean distance between camera poses extracted from predicted videos and those extracted from reference videos, where both sets of camera poses are obtained using GLOMAP~\cite{glomap}.
(3) \textbf{CLIP}: computes feature similarity in the CLIP~\cite{clip} space to evaluate semantic relevance to the given text prompt; 
(4) \textbf{FID, FVD}: measures distribution differences between the generated videos and the reference, providing an assessment of generation quality.

\vspace{1pt}\noindent\textbf{\ding{113} Comparison with Related Works.} 
As shown in Tab.~\ref{tab:compare_works}, given the diverse sources of game datasets, varying video resolutions, and different levels of control across methods, establishing a unified benchmark for comparison is challenging. Regarding our main contribution of scene-generalizable action control, the only capable technical paper~\cite{matrix} validated their approach on a simple action space (left turn, right turn, acceleration in racing games). In contrast, we validate our method on a much more comprehensive action space (forward/backward/left/right movement, jumping, acceleration/deceleration, mouse movement) and provide extensive results in the supplementary materials.

\subsection{Action Controllability}
\vspace{1pt}\noindent\textbf{\ding{113} Ablation Study.}
\begin{figure}[t]
  \centering
%   \vspace{-0.1cm}
  \includegraphics[width=1\linewidth]{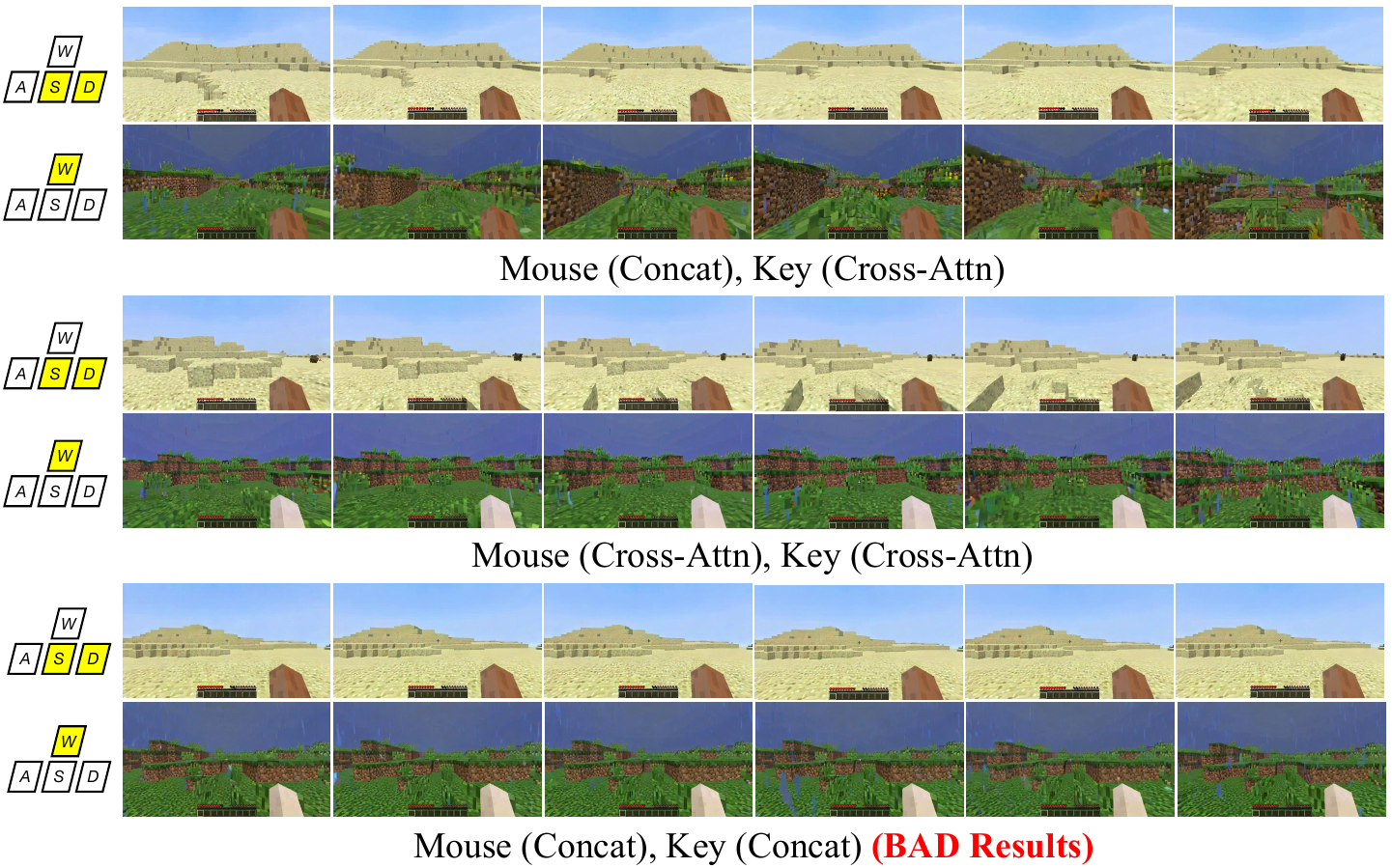}
  % \vspace{-0.6cm}
  \caption{Qualitative comparison of key input control performance. 
  % It can be observed that cross-attention significantly outperforms concatenation in handling discrete key input signals, while concatenation may fail to respond to the key input. 
  The yellow buttons indicate \textbf{pressed keys}.
  }
  \vspace{-0.2cm}
\label{fig:acm_ablation} 
\end{figure}
\begin{table}[t]\footnotesize

\vspace{-0.2cm}
\tabcolsep=0.09cm
\center
\begin{tabular}{c c | c c c c c c }
\toprule
  Strategy & Domain  & Cam$\downarrow$ & Flow$\downarrow$ & Dom$\uparrow$ & CLIP$\uparrow$ & FID$\downarrow$ & FVD$\downarrow$ \\
  \hline
  \hline
  Multi-Phase & In- & \textbf{0.0839} & \textbf{43.48} & - & - & - & -\\
  Multi-Phase & Open- & 0.0997 & 54.13 & \textbf{0.7565} & \textbf{0.3181} & \textbf{121.18} & \textbf{1256.94} \\
  One-Phase & Open- & 0.1134 & 76.02 & 0.7345 & 0.3111 & 167.79 & 1323.58 \\

\bottomrule

\end{tabular}
\caption{Quantitative results of evaluation on scene generalization.}
\vspace{-0.5cm}
\label{table:eval_scene_general}
\end{table}
\begin{figure}[t]
  \centering
%   \vspace{-0.1cm}
  \includegraphics[width=1\linewidth]{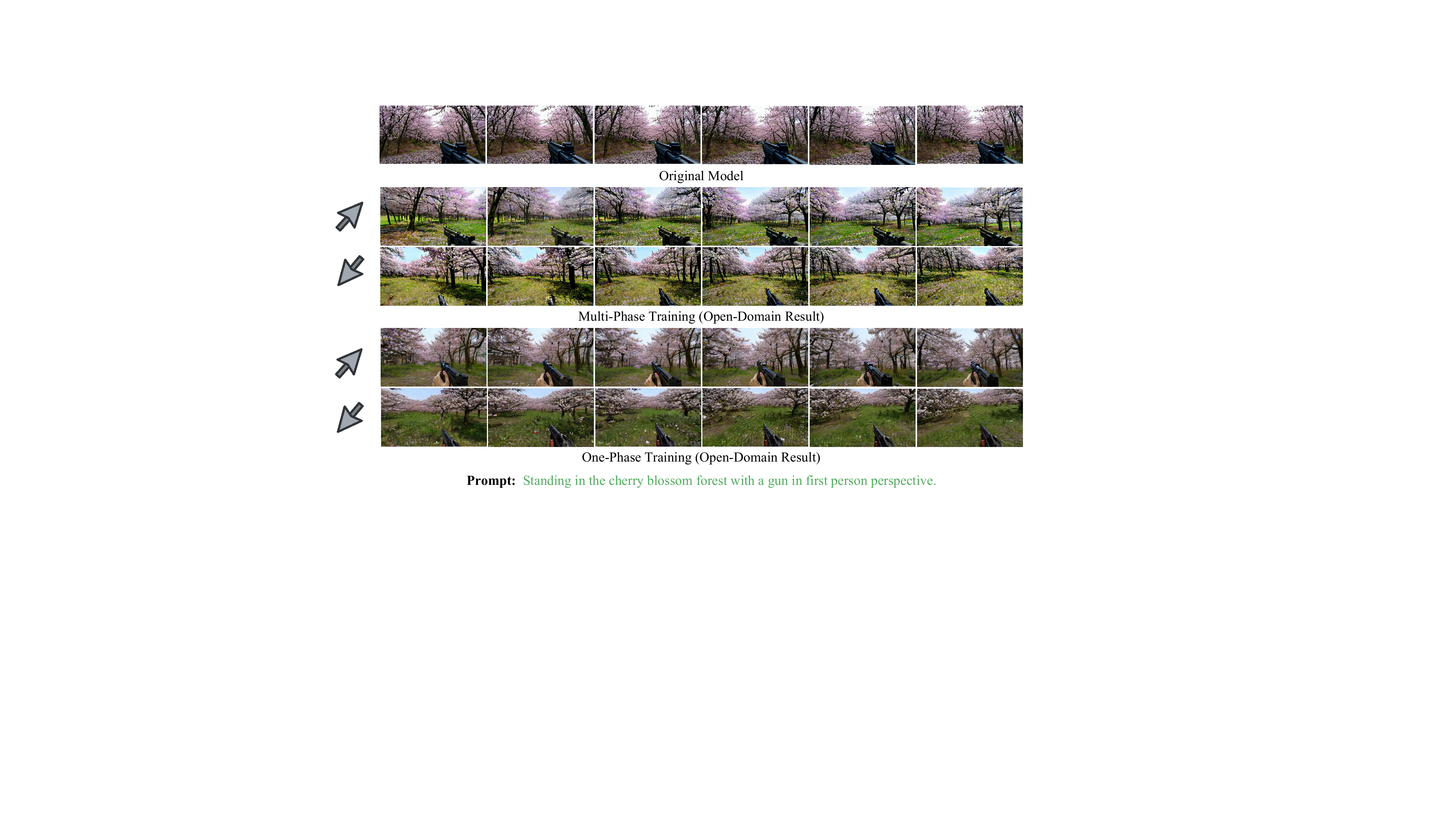}
  % \vspace{-0.6cm}
  \caption{Qualitative comparison of multi-phase training with one-phase training for scene generalization. The arrows represent the direction of \textbf{mouse movements}.
  }
  \vspace{-0.5cm}
\label{fig:comp_scene_general} 
\end{figure}
We perform ablation studies on the control mechanisms for continuous mouse movement and discrete keyboard control, comparing two typical methods: cross-attention and concatenation. The results are presented in Tab.~\ref{table:ablation} and Fig.~\ref{fig:acm_ablation}. 
For discrete keyboard inputs, cross-attention outperforms concatenation. As shown in Fig.~\ref{fig:acm_ablation},  concatenating the action control signal with the input leads to poor action following performance. This suggests that category-based signal control benefits from similarity-based cross-attention, as is often used in text-based control.
In contrast, for continuous mouse movement, concatenation is more effective than cross-attention. This may be due to cross-attention’s similarity computation, which tends to reduce the influence of the control signal’s magnitude, thereby affecting the final result. 
Additionally, the values of \textbf{Cam} and \textbf{Flow} reveal that mouse movements have a larger impact on the visual output than keyboard inputs, particularly in the mouse-large test set where movement magnitude is greater.
For the metrics of \textbf{CLIP}, \textbf{FID} and \textbf{FVD}, there is little difference between the methods. This is primarily because our multi-phase training strategy decouples visual style learning into \textbf{Phase \#1}, where style learning is consistent across all methods. As a result, metrics assessing semantic consistency and generation quality yield similar performance across different methods.

\subsection{Scene Generalization}
\vspace{1pt}\noindent\textbf{\ding{113} Create new games in generalized scenes.}
In Fig.~\ref{fig:overview}, we showcase a variety of newly generated game videos across open-domain scenes. These results inspire a future where generative game engines emerge as a new form of gaming, allowing players or game creators to generate and interact with anything they can imagine at minimal cost. 

\vspace{1pt}\noindent\textbf{\ding{113} Comparison Results.}
As shown in Tab.~\ref{table:eval_scene_general}, we evaluate the action following capability using \textbf{Cam} and \textbf{Flow} metrics on in-domain as baseline. The results demonstrate that our multi-phase training strategy achieves better action following performance closer to the baseline compared to the one-phase approach. Furthermore, the multi-phase strategy shows superior performance across text-alignment and generation quality metrics including \textbf{CLIP}, \textbf{FID}, and \textbf{FVD}. The \textbf{Dom} metric, which measures the CLIP space similarity between videos generated by the original model and the fine-tuned model, indicates that the multi-phase trained model maintains a domain closer to the original model. This is visually confirmed in Fig.~\ref{fig:comp_scene_general}, where the multi-phase strategy preserves the original model's generation domain without style leakage. Additionally, the one-phase approach degrades the generated video quality with noticeable artifacts.

% \textbf{More results on scene generalization and the evaluation of scene generalization capabilities can be found in the supplementary materials.}

% \vspace{1pt}\noindent\textbf{\ding{113} More inspiration from examples of racing games.}
% In Fig.~\ref{fig:discussion}, we discuss an intriguing example. Since our collected Minecraft data is from a first-person perspective, the learned action space primarily generalizes to first-person scenes. Here, we experimented with a racing game scenario using a prompt for a car. Interestingly, we observed that the model’s learned yaw control for the mouse seamlessly generalized to steering control in the racing game. Additionally, certain directional controls, such as moving backward or sideways, were diminished, an adaptation that aligns well with typical controls in a racing game, where these actions are rarely needed. This example not only highlights the strong generalization capabilities of our method but also raises the question: could there exist a larger, more versatile action space that encompasses a wider range of game controls, extending beyond first-person and racing games?

% This example also leads us to wonder whether the racing game scenario could have applications in autonomous driving. If we were to collect an action dataset from an autonomous driving simulation environment, could a pre-trained model then generate unlimited open-domain autonomous driving data? Our exploration of scene generalization within generative game engines may hold valuable insights for other fields as well.

\subsection{Evaluation for GF-Minecraft Dataset}
\begin{table}[t]\footnotesize

\vspace{-0.2cm}
\tabcolsep=0.09cm
\center
\begin{tabular}{c | c c c c c }
\toprule
  Dataset  & Cam$\downarrow$ & Flow$\downarrow$ & CLIP$\uparrow$ & FID$\downarrow$ & FVD$\downarrow$ \\
  \hline
  \hline
  VPT~\cite{vpt} & 0.1324 & 107.67 & \textbf{0.3174} & 156.69 & 1233.15  \\
  GF-Minecraft (ours) & \textbf{0.0839} & \textbf{43.48} & 0.3135 & \textbf{125.85} & \textbf{1047.59} \\

\bottomrule

\end{tabular}
\caption{Comparison with Minecraft dataset with human bias.}
\vspace{-0.2cm}
\label{table:comp_dataset}
\end{table}
\begin{table}[t]\footnotesize

\vspace{-0.2cm}
\tabcolsep=0.09cm
\center
\begin{tabular}{c | c c c c c c c }
\toprule
  Dataset  & W & A & S & D & Space & Shift & Ctrl \\
  \hline\hline
  VPT~\cite{vpt} & 50.11\% & 4.03\% & 0.32\% & 3.45\% & 20.37\% & 0.14\% & 19.58\%   \\
  Ours & 13.56\% & 13.56\% & 13.56\% & 13.56\% & 15.25\% & 15.25\% & 15.25\%  \\

\bottomrule

\end{tabular}
\caption{The proportion of key inputs across different datasets.}
\vspace{-0.2cm}
\label{table:summary_dataset_freq}
\end{table}
\begin{figure}[t]
  \centering
%   \vspace{-0.1cm}
  \includegraphics[width=1\linewidth]{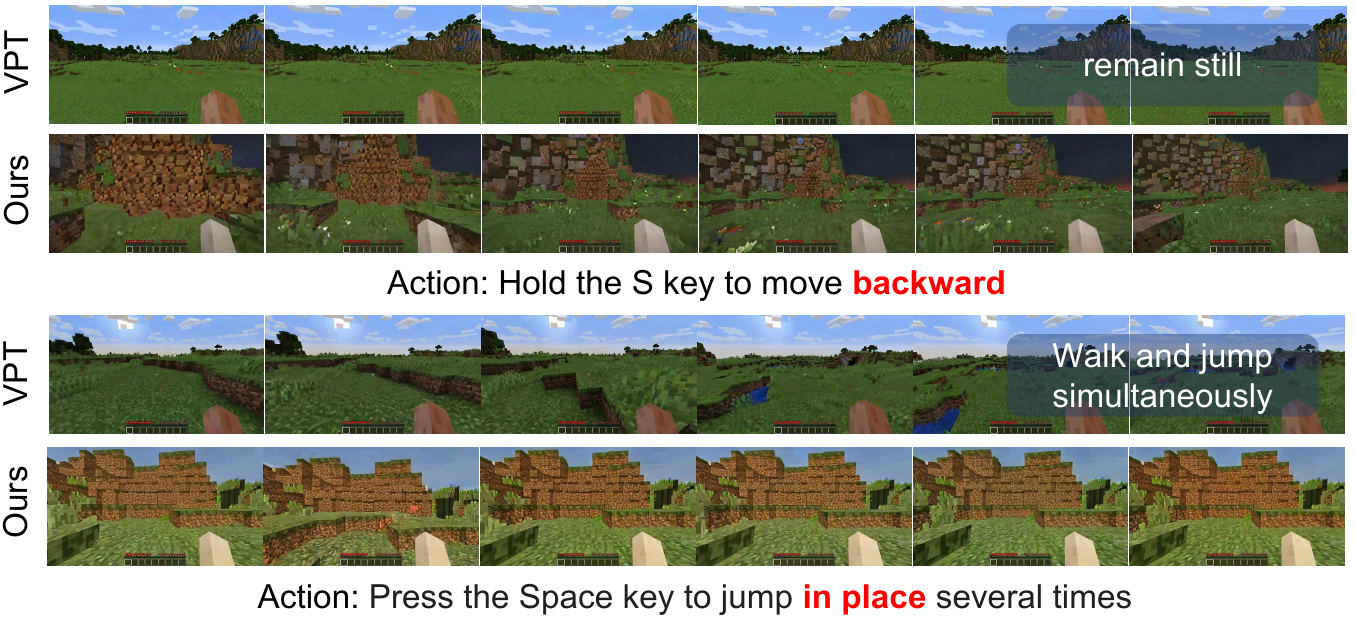}
  % \vspace{-0.6cm}
  \caption{
  Compare the dataset on actions that are less commonly used by human players to test the effect of human bias in dataset.
  }
\vspace{-0.5cm}
\label{fig:comp_dataset} 
\end{figure}
In Sec.~\ref{subsec:data_collect}, we introduced our GF-Minecraft dataset, which eliminates human bias by ensuring uniform distribution of atomic actions. This design enables models trained on GF-Minecraft to effectively respond to actions that human players rarely perform. We compare our dataset with the VPT~\cite{vpt} dataset, which contains recordings of human gameplay and inherent human biases. Specifically, we selected video clips from VPT's Find Cave dataset, as it closely matches our setting by primarily excluding inventory management and block modification operations.
Tab.~\ref{table:summary_dataset_freq} compares the usage proportions of keyboard inputs across both datasets. The VPT dataset shows highly skewed distributions. For instance, the forward movement key (W) appears over 100 times more frequently than the backward movement key (S), reflecting typical human gameplay patterns. Tab.~\ref{table:comp_dataset} compares the action control capabilities of models trained on both datasets through in-domain evaluation, demonstrating clear advantages of our GF-Minecraft dataset in action following performance.
Fig.~\ref{fig:comp_dataset} illustrates two representative examples: jumping in place and moving backward (both actions rarely performed by human players). The model trained on GF-Minecraft successfully follows these actions, while the VPT-trained model fails to execute them properly. Specifically, when commanded to jump in place, the VPT model incorrectly combines forward movement with jumping, and when instructed to move backward, it simply remains stationary.

\subsection{Evaluation for Long Video Generation}
\begin{table}[t]\footnotesize

\vspace{-0.2cm}
\tabcolsep=0.09cm
\center
\begin{tabular}{c | c c c c c }
\toprule
  Loss Scope  & Cam$\downarrow$ & Flow$\downarrow$ & CLIP$\uparrow$ & FID$\downarrow$ & FVD$\downarrow$ \\
  \hline
  \hline
  All frames & 0.1547 & 148.73 & 0.2965 & 176.07 & 1592.43 \\
 Only predicted frames & \textbf{0.0924} & \textbf{85.45} & \textbf{0.3190} & \textbf{136.95} & \textbf{1154.45} \\

\bottomrule

\end{tabular}
\caption{Ablation study on loss scope for long video generation.}
\vspace{-0.2cm}
\label{table:abalation_loss}
\end{table}
\begin{figure}[t]
  \centering
%   \vspace{-0.1cm}
  \includegraphics[width=1\linewidth]{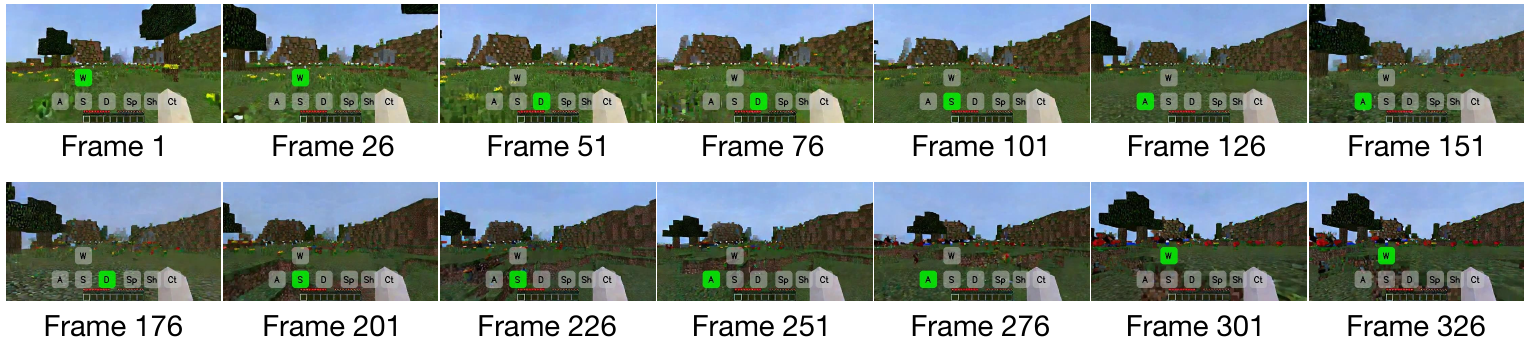}
  % \vspace{-0.6cm}
  \caption{
  Demonstration of key frames in generated long video. 
  }
\vspace{-0.5cm}
\label{fig:long_video} 
\end{figure}
In Tab.~\ref{table:abalation_loss}, we compare different scopes for computing loss functions in long video generation training: calculating losses across all frames versus only on frames that need to be predicted. The results demonstrate that computing losses solely on frames requiring prediction yields better performance. This improvement can be attributed to the elimination of noise interference from previously generated frames, as learning from such noise is irrelevant for future video generation.
As demonstrated in Fig.~\ref{fig:ar_gen}, our model successfully generates long videos exceeding 300 frames in length. Additional examples of long video generation can be found in the supplementary materials.

\section{Conclusion}
% In this paper, we propose GameFactory using generative interactive videos to create new games, addressing gaps in related research from the perspectives of action control and scene generalization.
% We propose an action control module that can manage continuous mouse movements and discrete keyboard inputs, as well as a multi-phase training mechanism to achieve scene generalization.
% We also introduce a dataset based on the Minecraft game, which includes various action combinations, rich scenes, and text annotations.
% In the future, we plan to extend the boundaries of generative game engines based on this work, aiming to create brand new games that players can interactively play.
In this paper, we propose GameFactory, a framework that uses generative interactive videos to create new games, addressing significant gaps in existing research, particularly in scene generalization. 
% \jiwen{\sout{While an ideal, fully playable generative game engine would require auto-regressive video generation, no pretrained powerful auto-regressive models are currently available. Therefore, we begin our exploration using a pretrained full-sequence text-to-video model in this work, and our approach can naturally extend to powerful future auto-regressive models, such as those incorporating diffusion forcing~\cite{diffusion-forcing}.}}
The study of generative game engines still faces many challenges, including the design of diverse levels and gameplay, player feedback systems, in-game object manipulation, long-context memory, and real-time game generation. GameFactory marks our first effort in this field, and we aim to continue progressing toward the realization of a fully capable generative game engine.

% \section*{Acknowledgements}
% This work is partially supported by the National Nature Science Foundation of China (No. 62402406).

{
    \small
    \bibliographystyle{ieeenat_fullname}
    \bibliography{main}
}

\appendix
\onecolumn
\begin{appendices}

% Additional results of action control in Minecraft and open-domain scenarios can be found in the attached HTML file.
Additional results of action control in Minecraft and open-domain scenarios can be found in our homepage at \url{https://yujiwen.github.io/gamefactory/}.
\section{Details of Minecraft Dataset}

\subsection{Basic Information}
For data collection, we use Minedojo~\cite{minedojo} to obtain Minecraft snapshots which contain a wide array of details within the agent’s surroundings. To mitigate biases introduced by human player habits, we uniformly sampled the frequency and duration of each action and randomly combined them to enhance action generalization (e.g., pressing two keys simultaneously or combining keyboard and mouse operations). Subsequently, the agent was connected to the MineDojo simulation environment to execute the action sequences and collect corresponding observations. 

To further enhance the diversity and generalization of the generated game scenes, we preconfigured three biomes (forest, plains, desert), three weather conditions (clear, rain, thunder), and six different times of day ("Starting of a day," "Noon, sun is at its peak," "Sunset," "Beginning of night," "Midnight, moon is at its peak," "Beginning of sunrise"). This process resulted in the generation of 2,000 video clips with action annotations, each containing 2,000 frames.

\subsection{Data PreProcessing}
\begin{figure*}[!htbp]
  \centering
%   \vspace{-0.1cm}
  \includegraphics[width=1\linewidth]{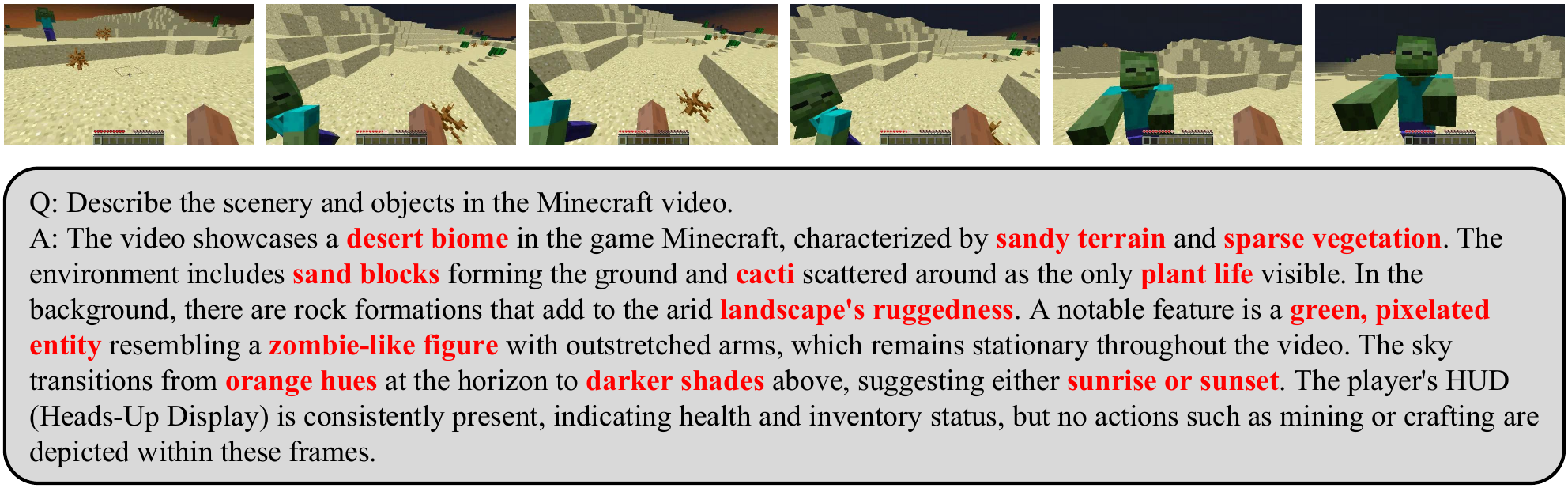}
  % \vspace{-0.6cm}
  \caption{An example of video clip annotation, where words describing scenes and objects are highlighted in red and bolded.
  }
  \vspace{-0.2cm}
\label{fig:annotation} 
\end{figure*}

After obtaining the full video, we preprocess it in two steps:
(1) \textbf{Slicing:} Randomly sample consecutive sequences of $k$ frames from the full video to form new video clips. Sampling continues until the total number of sampled frames reaches $n$ times the total frame count of the original video. In our experimental setup, $k=81$, which is slightly larger than the generation length of the video diffusion model ($77$), and $n=3$; 
(2) \textbf{Text Annotation:} We use the open-source multimodal large language model MiniCPM-V~\cite{minicpm} to annotate the sliced video clips with text. The corresponding prompts and annotation examples are shown in Fig.~\ref{fig:annotation}.

\subsection{Details of Action Space}
We use the part of the action space of Minedojo~\cite{minedojo} which encompasses nearly all actions available to human players, including keypresses, mouse movements, and clicks. We used keypresses and mouse movements as our control signal. The specific binary actions used in our setup are listed in Tab~\ref{sup_mc_as}. Interface$_1$ to Interface$_5$ represent different MineDojo interfaces, where mutually exclusive actions (e.g., moving forward and backward) are assigned to the same interface. Mouse movements are represented by the offset of the mouse pointer from the center of the game region. For each frame in the video, we calculate the cumulative offset relative to the first frame, and this absolute offset is used as input to the model.

\begin{table}[t]
\centering
\caption{\textbf{Details of Action Space.} The term \textit{Control Signal} refers to the raw input signals utilized for training purposes, while the \textit{Action Interface} represents the corresponding interface in the MineDojo platform that maps these input signals to actionable commands.}
\begin{tabular}{|l|l|l|}
\toprule
\textbf{Behavior}   & \textbf{Control Signal}   & \textbf{Action Interface}\\ \hline
forward   & W key        & Interface$_1$   \\ \hline
back      & S key        & Interface$_1$   \\ \hline
left      & A key        & Interface$_2$   \\ \hline
right     & D key        & Interface$_2$   \\ \hline
jump      & space key    & Interface$_3$   \\ \hline
sneak     & shift key    & Interface$_3$   \\ \hline
sprint    & ctrl key     & Interface$_3$   \\ \hline
vertical perspective movement    & mouse movement(yaw)     & Interface$_4$   \\ \hline 
horizontal perspective movement    & mouse movement(pitch)     & Interface$_5$   \\ 
\bottomrule
\end{tabular}
\label{sup_mc_as}
\end{table}

% It is worth noting some unique physical rules within the Minecraft domain that result in distinct visual effects:

% \begin{itemize}
%     \item Holding \textbf{Ctrl} to sprint increases the field of view (FOV) and boosts movement speed by approximately 30\%. However, this effect is limited to forward movement and cannot be combined with other directional inputs.
%     \item Holding \textbf{Shift} reduces movement speed and can be combined with any directional key, allowing for slowed movement in any direction.
% \end{itemize}
\section{Supplementary Experimental Results}

\subsection{Interaction in generative Minecraft videos.}
 \begin{figure}[!tbp]
  \centering
  % \vspace{-0.1cm}
  \includegraphics[width=1\linewidth]{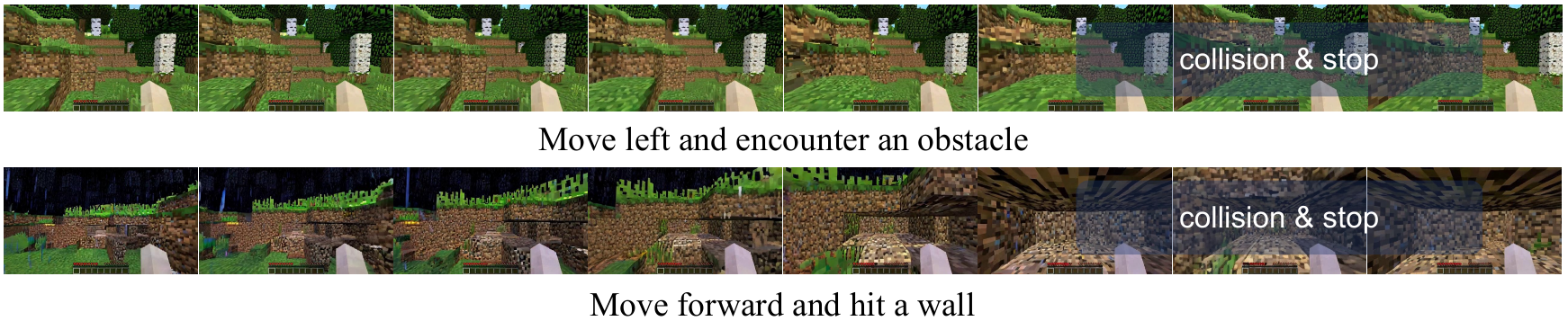}
  % \vspace{-0.7cm}
  \caption{Demonstration of the learned response to collision, one of the most common interactions in Minecraft navigation. Note that the text below each video frame is a descriptive label of the content, not a text prompt provided to the model.}
  \vspace{-0.1cm}
\label{fig:mc_interact} 
\end{figure}
For navigation agents, collision is one of the most critical physical interactions in a simulator. Since our data collection process involves randomly generated scenes, our dataset naturally includes numerous examples of collisions. In these cases, even when action inputs are provided, the agent's behavior should ideally remain stationary. Such corner cases can impact the learning of the action control module, especially when data volume is limited. However, during inference, we observed that our model has developed collision detection ability and provides appropriate interaction feedback, as shown in Fig.~\ref{fig:mc_interact}.

\subsection{More Inspiration from Examples of Racing Games}
\begin{figure*}[!htbp]
  \centering
%   \vspace{-0.1cm}
  \includegraphics[width=1\linewidth]{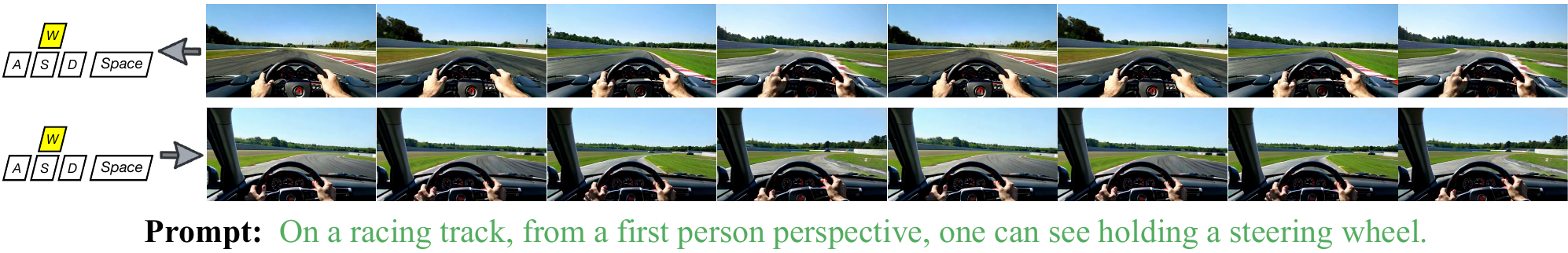}
  % \vspace{-0.6cm}
  \caption{Our model demonstrates the ability to generalize to a different game type, a racing game. Interestingly, the yaw control learned in Minecraft seamlessly transfers to steering control in the racing game, while unrelated actions, such as moving backward, left, or right, and pitch angle adjustments, automatically diminish.
  }
  \vspace{-0.2cm}
\label{fig:discussion} 
\end{figure*}

In Fig.~\ref{fig:discussion}, we discuss an intriguing example. Since our collected Minecraft data is from a first-person perspective, the learned action space primarily generalizes to first-person scenes. Here, we experimented with a racing game scenario using a prompt for a car. Interestingly, we observed that the model’s learned yaw control for the mouse seamlessly generalized to steering control in the racing game. Additionally, certain directional controls, such as moving backward or sideways, were diminished, an adaptation that aligns well with typical controls in a racing game, where these actions are rarely needed. This example not only highlights the strong generalization capabilities of our method but also raises the question: could there exist a larger, more versatile action space that encompasses a wider range of game controls, extending beyond first-person and racing games?
This example also leads us to wonder whether the racing game scenario could have applications in autonomous driving. If we were to collect an action dataset from an autonomous driving simulation environment, could a pre-trained model then generate unlimited open-domain autonomous driving data? Our exploration of scene generalization within generative game engines may hold valuable insights for other fields as well.

%  \begin{figure}[!tbp]
%   \centering
%   % \vspace{-0.1cm}
%   \includegraphics[width=1\linewidth]{figure/discussion.pdf}
%   % \vspace{-0.7cm}
%   \caption{Our model demonstrates the ability to generalize to a different game type, a racing game. Interestingly, the yaw control learned in Minecraft seamlessly transfers to steering control in the racing game, while unrelated actions automatically diminish.}
%   % \vspace{-0.1cm}
% \label{fig:discussion} 
% \end{figure}

% \section{Implementation Details of Autoregressive Long Video Generation} 
% Since the model structure remains unchanged and only the noise levels vary across frames, the training configuration remains identical to full-sequence generation, including action control. We also add small noise perturbations (equivalent to step $15$/$1000$) to conditional frames, which helps reduce error accumulation in long-term generation~\cite{gamengine, oasis}.
% The long video results for action control can be found in the attached HTML file.

\section{Potential of Generalizable World Model}
We propose that the \textbf{GameFactory} we have developed is not merely a tool for creating new games but a \textbf{Generalizable World Model} with far-reaching implications. This model has the capability to generalize physical knowledge learned from small-scale labeled datasets to open-domain scenarios, addressing challenges in areas like autonomous driving~\cite{gao2024vista,hu2023gaia} and embodied AI~\cite{zhu2024irasim,bar2024navigation,qin2024worldsimbench,yang2023learning}, which also face limitations due to the lack of large-scale action-labeled datasets.
The generalizable world model has two key applications from different perspectives:
\begin{itemize}
    \item \textbf{As a data producer:} It transfers knowledge from small labeled datasets to open-domain scenarios, enabling the generation of diverse unlimited action-annotated data that closely approximates real-world complexity.
    \item \textbf{As a simulator:} It provides an environment for directly training agents to perform real-world tasks~\cite{ajay2024compositional,qin2024mp5,huang2023embodied}, closely approximating real-world conditions. By enabling controlled and diverse scenario generation, including extreme situations that are difficult to capture in real-world data collection, it facilitates the development of policy models that are exposed to a wide range of environments and interactions, thereby improving their robustness and generalizability, and aiding in overcoming the challenges of sim-to-real transfer.
\end{itemize}

\end{appendices}

\end{document}